\documentclass{article}
\usepackage{amssymb}
\usepackage{hyperref}

\usepackage[preprint]{corl_2025} 

\usepackage{moreverb,url}
\usepackage{xspace}
\usepackage{xcolor}

\usepackage{amssymb}
\usepackage{amsmath}
\usepackage{amsthm}
\usepackage{amsbsy}
\usepackage{bbm}
\usepackage{dsfont}

\usepackage{graphicx}
\usepackage{wrapfig}
\usepackage{booktabs} 

\usepackage{subfiles}
\usepackage{csquotes}

\usepackage{array}

\usepackage{graphicx}
\usepackage{multirow}
\usepackage[caption=false]{subfig}

\newcommand\BibTeX{{\rmfamily B\kern-.05em \textsc{i\kern-.025em b}\kern-.08em
T\kern-.1667em\lower.7ex\hbox{E}\kern-.125emX}}

\usepackage[dvipsnames]{xcolor} 
\newcommand{\commint}[2][red]{\textcolor{#1}{\textbf{#2}}}

\definecolor{wine}{RGB}{204, 0, 102}
\definecolor{magenta_wine}{RGB}{158, 44, 143}
\definecolor{dusty_wine}{RGB}{143, 59, 101}
\definecolor{ocean}{RGB}{13, 121, 202}
\definecolor{light_ocean}{RGB}{18, 178, 235}
\definecolor{dark_ocean}{RGB}{10, 89, 148}
\definecolor{grey}{RGB}{170, 170, 170}
\definecolor{light-grey}{RGB}{220, 220, 220}
\definecolor{dark_gray}{rgb}{0.2, 0.2, 0.2} 
\definecolor{med-grey}{rgb}{0.3, 0.3, 0.3} 
\definecolor{grape}{RGB}{112,48,160}
\definecolor{aqua}{RGB}{52,172,139}
\definecolor{dark_aqua}{RGB}{35,115,93}
\definecolor{dark_orange}{RGB}{216,92,0}
\definecolor{vibrant_orange}{RGB}{250, 160, 26}
\definecolor{vibrant_blue}{RGB}{14, 120, 255}
\definecolor{vibrant_pink}{RGB}{255, 0, 104}
\definecolor{dark_red}{RGB}{122, 0, 0}
\definecolor{dark_green}{RGB}{0, 92, 34}
\definecolor{dusty_blue}{RGB}{77, 91, 128}
\definecolor{dark_brown}{RGB}{125, 54, 36}

\theoremstyle{subtle} 

\newcounter{qnum}
\newcommand{\deltaarrays}{{delta arrays}\xspace}
\newcommand{\DDM}{{\textbf{DDM}}\xspace}
\newcommand{\matsac}{{\textit{MATSAC}}\xspace}
\newcommand{\matbc}{{\textit{MATBC}}\xspace}
\newcommand{\matbcft}{{\textit{MATBC-FT}}\xspace}
\newcommand{\matbcftog}{{\textit{MATBC-FT-OG}}\xspace}
\newcommand{\matbcftdec}{{\textit{MATBC-FT-DEC}}\xspace}
\newcommand{\matbcftcec}{{\textit{MATBC-FT-CEC}}\xspace}
\newcommand{\matbcftgec}{{\textit{MATBC-FT-MEC}}\xspace}

\newcommand{\lcrope}{{\textit{LCRoPE}}\xspace}
\newcommand{\lre}{{\textit{LRE}}\xspace}
\newcommand{\learned}{{\textit{LE}}\xspace}

\newcommand{\asel}{\mathcal{A}_{\text{SEL}_\text{xy}}}
\newcommand{\Nsel}{\N_{\textsc{SEL}}}

\newcommand{\rog}{r_{\textsc{OG}}}
\newcommand{\rdec}{r_{\textsc{DEC}}}
\newcommand{\rcec}{r_{\textsc{CEC}}}
\newcommand{\rgec}{r_{\textsc{GEC}}}

\usepackage{fontawesome5}

\title{Distributed Dexterous Manipulation with Spatially Conditioned Multi-Agent Transformers}

\author{%
  \centerline{Sarvesh Patil}\\
  \centerline{\texttt{sarveshp@andrew.cmu.edu}}\\
  \centerline{Carnegie Mellon University, The Robotics Institute}\\
  \centerline{Pittsburgh, PA, United States}
}

\begin{document}

\maketitle

\begin{abstract}
Distributed Dexterous Manipulation (DDM) is a novel paradigm that presents significant control challenges due to high action-space redundancy, inter-robot cooperation, and dynamic object-robot interactions. This paper introduces a framework based on spatially conditioned Multi-Agent Transformers (MATs) to efficiently learn robust control policies for a DDM system grounded in an array of 64 soft delta robots arranged in an $8\times8$ grid. 
Our three core contributions are: (i) an MAT with adaptive layer norm for compute efficiency, (ii) spatial contrastive embeddings to ground transformer embeddings in the spatial configuration of the robots, and (iii) an MAT-based behavior cloning method fine-tuned using Soft Actor Critic. We also propose an action selection formulation to analyze the trade-off between task performance and the number of robots utilized. 
Our experiments show that MATs iteratively refine their actions through the stacked attention blocks. This further informs the benefit of spatial conditioning in transformers to learn DDM policies. We demonstrate long-horizon planar manipulation tasks with objects of various geometries in simulation and real-world. Finally, we show how action selection mitigates robot maintenance by reducing wear and tear due to inter-robot collisions while maintaining the ability to manipulate objects along various trajectories in the real-world, achieving an average error of $\sim1.5$ cm, while using $\sim65\%$ fewer robots. Website and Code can be found 
\url{https://distributed-dexterous-manipulation.github.io/}
and 
\url{https://github.com/Servo97/dexterous-manipulation-delta-arrays}
\end{abstract}



\newcommand{\sarvesh}[1]{\commint[RedOrange]{[Sarvesh: #1]}}
\newcommand{\zeynep}[1]{\commint[TealBlue]{[Zeynep: #1]}}
\newcommand{\oliver}[1]{\commint[WildStrawberry]{[Oliver: #1]}}
\newcommand{\sidney}[1]{\commint[Orange]{[Sidney: #1]}}


\section{Introduction}
\begin{figure}[t!]
    \centering
    \includegraphics[width=0.99\textwidth]{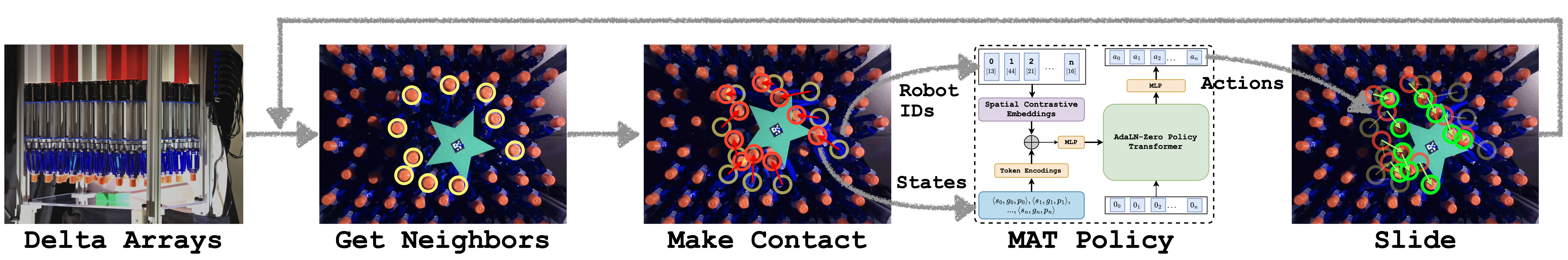}
    \caption{We present Distributed Dexterous Manipulation as a novel paradigm for cooperative manipulation using arrays of soft delta robots. We propose a spatially grounded Multi-Agent Transformer to train sample-efficient policies using off-policy reinforcement learning and behavior cloning.}
    \label{fig:main_fig}
    \vspace{-0.5cm}
\end{figure}

Enabling robots to dexterously manipulate a wide variety of objects remains a fundamental challenge. Distributed manipulation systems, where multiple actuators impart desired motion on an object through a combination of external forces,
offer a promising approach.  Examples include smart conveyors \citep{howie1, howie2} and linear actuator arrays \citep{mit, MIT_inFORM} that move objects through coordinated patterns. Recently, \citet{delta_array} introduced \deltaarrays, a distributed dexterous manipulation (\DDM) system comprising an $8\times8$ array of 3-DoF compliant delta robots as shown in Fig. \ref{fig:main_fig}. This system can apply forces dynamically by engaging varying groups of robots, allowing for dexterous manipulation of diverse objects.

Learning effective control policies for \deltaarrays is particularly challenging due to: (1) the extreme redundancy in the collective action space (192-DoF) resulting in highly multi-modal dynamics; (2) the need for robots to communicate intent and cooperate within dynamically changing neighborhoods based on object state; and (3) the requirement for sample-efficient learning methods suitable for complex, contact-rich interactions. However, controlling such \DDM systems presents significant hurdles. Prior works have focused on rotary actuator arrays \citep{murphey2003experiments, howie1} or linear actuators \citep{tsinghua}, limiting their applicability to the full 3-DoF capabilities needed for dexterous tasks. 

To address these challenges, we propose a learning framework centered around spatially conditioned Multi-Agent Transformers (MATs) for manipulating objects on an $\mathbb{XY}$ plane. Transformers \cite{vaswani} are well-suited for \DDM due to their ability to handle robot sequences of varying lengths (dynamic neighborhoods) and learn effective representations between objects and robots. To explicitly ground the attention layers in MATs to the physical layout of \deltaarrays, we introduce Spatial Contrastive Embeddings (SCEs), which encode the spatial arrangement of the robots in the array. Further, to address sample efficiency, we employ a two-phase learning strategy: (i) pre-training the MAT policy via Behavior Cloning (\matbc) using demonstrations from a visual servoing expert, (ii) fine-tuning with off-policy Reinforcement Learning using Soft Actor-Critic (\matsac) \citep{SAC}. We refer to this two-phase fine-tuned policy as \matbcft. Unlike the hand-engineered visual servoing policy, which follows fixed flow vectors, \matbcft learns to anticipate object dynamics and adaptively coordinates robots, resulting in smoother trajectories and more robust manipulation across various object geometries. Further, we experiment with reward shaping to affect the $\mathbb{Z}$ component of the action space to select or deselect robots. 
This induces the policy to learn an optimal trade-off between accurately manipulating objects and selecting the number of agents. This also mitigates physical wear and prevents entanglement of the compliant delta linkages on the real hardware.

The core contributions of this paper are:
\begin{enumerate}
    \item A framework using Spatially Conditioned Multi-Agent Transformers combined with Behavior Cloning and Reinforcement Learning Fine-tuning (\matbcft) for learning sample-efficient and robust policies for complex \DDM tasks
    \vspace{-0.1cm} 
    \item Spatial Contrastive Embeddings (SCEs) as a novel method to explicitly encode spatial relationships between robots, enhancing coordination within the MAT for \DDM.
    \vspace{-0.1cm} 
    \item An analysis of reward shaping mechanisms for efficient robot utilization by learning a trade-off between task success and the number of active agents.
    \vspace{-0.1cm} 
    \item Providing empirical validation and analysis on the benefits of iterative refinement from message-passing through transformer attention layers for cooperative manipulation tasks. 
\end{enumerate}

\section{Related Work}
\label{sec:rel_wrk}


\textbf{Distributed Manipulation}: Distributed manipulation systems employ multiple actuators, often in 2D tessellating structures, for coordinated object motions. Implementations range from 
air-jet arrays \cite{airjets} to surfaces with actuated joints \cite{mem_arrays, howie1, howie2} and actuated workbenches \cite{actuated_workbench, 2.5d_display}. While some works have explored 3D cooperative pushing \cite{rus1997dist, rus1997minimalism, rus2000}, these systems are limited by the end-effector dexterity and workspace constraints \cite{tsinghua}. The \deltaarrays system \cite{delta_array}, used in our work, represents a step towards \DDM by offering a dense array of 3-DoF compliant delta robots. They enable complex interactions but also introduce significant control challenges due to the high dimensionality of their action spaces and the need for coordination within the shared workspace of the robots.


\textbf{Behavior Cloning (BC) and Reinforcement Learning (RL) for Dexterous Manipulation:} BC learns policies by imitating expert demonstrations \cite{bc1, bc2, decision_tf, bczero}, while RL develops strategies through trial-and-error \cite{levine2016end}. RL has shown success in complex single-agent dexterous manipulation \cite{openai, rajeswaran2017learning, popov2017data} using both on-policy (e.g., Proximal Policy Optimization (PPO) \cite{ppo}) and off-policy (e.g., Twin Delayed Deep Deterministic Policy Gradient (TD3) \cite{td3} Soft Actor-Critic (SAC) \cite{SAC}) methods. However, RL often suffers from poor sample efficiency, requiring extensive interaction \cite{kalashnikov18a}, which is problematic for real robots. Combining BC pre-training with RL fine-tuning has been proposed in literature to learn robust control policies \cite{bc0, vikash1, real0, real1}. Offline RL has also been explored for multi-agent settings \cite{offlinemarl, vikash2}, but applying these methods effectively to the fine-grained manipulation tasks in \DDM remains unexplored. Our methods utilize the sequence modeling capacity of transformers to learn \DDM policies.

\textbf{Multi-Agent RL (MARL) for Dexterous Manipulation:} \DDM inherently involves multiple cooperating robots. MARL often employs the paradigm of Centralized Training with Decentralized Execution. Methods used in \citet{COMA, QMIX, QTRAN} learn global value functions and decompose into local policies. Effective credit assignment becomes crucial for these methods. Transformers, on the other hand, have been proposed \cite{chen2023bi, MAT} as a Centralized Training, Centralized Execution method. They have shown great ability to learn sequential representations and handle varying input lengths \cite{gato, decision_tf, q_transformer}. \citet{MAT} proposed an MAT with PPO for simulated bi-manual tasks in simulation. Our methods induce implicit spatial reasoning needed for multi-robot systems like the delta arrays and show zero-shot transfer from simulation to real-world. 
\section{Problem Formulation and Preliminaries}
\label{sec:bgnd}
\subsection{Delta Arrays for Distributed Dexterous Manipulation}
\label{subsec:delta_arrays}
\begin{figure}[h]
    \centering
    \subfloat[Grasping]{\includegraphics[width=0.3\textwidth]{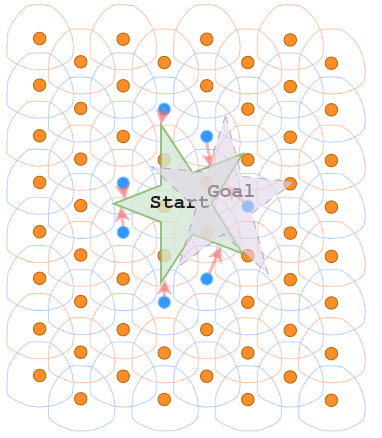}
    \label{fig:task_init}}
    \hfill
    \subfloat[Pushing]{\includegraphics[width=0.3\textwidth]{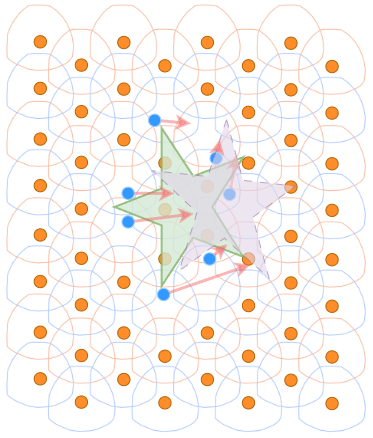}
    \label{fig:task_mid}}
    \hfill
    \subfloat[Final]{\includegraphics[width=0.3\textwidth]{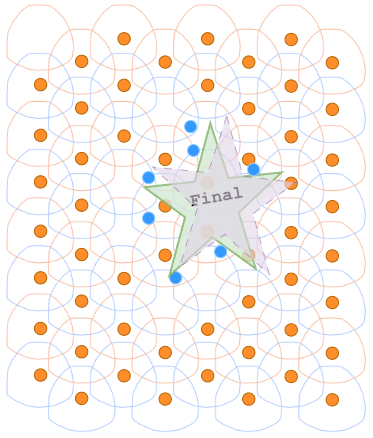}
    \label{fig:task_end}}
    \caption{}
    \label{fig:task}
\end{figure}

We perform our real-world evaluations on the delta arrays introduced by \citet{delta_array}. The array consists of 64 compliant delta robots, i.e., fingers, \cite{og_delta} inspired by the original delta mechanism \cite{clavel_delta_robot}. Each delta finger has three linear actuators, enabling control in the $\mathbb{XY}$ axes within an oval-shaped workspace of $\sim2.5$ cm radius, and $10$ cm for the $\mathbb{Z}$ axis. The result is a 192 DoF combined action space across the array. The delta robots are densely arranged in a hexagonal grid, with centers spaced $\sim4.35$ cm apart. This overlap of individual workspaces allows subsets of robots to collaborate on common tasks as well as large objects. 
We observe states using a bottom-up camera positioned beneath a glass plane that originates the world frame. The robot positions and workspaces are visualized in Fig. \ref{fig:task}. To accelerate training and testing, we implement a simulation environment in MuJoCo. Each delta robot is modeled as a floating capsule 3 DoF fingertip with real-world workspace constraints. 

\newcommand{\rb}{\mathcal{D}}
\newcommand{\bd}{\mathcal{B}}
\newcommand{\zlow}{a_{z}^\textsc{low}}
\newcommand{\zhigh}{a_{z}^\textsc{high}}
\newcommand{\B}{\mathcal{B}}

\newcommand{\N}{\mathcal{N}}
\newcommand{\St}{\mathcal{S}}
\newcommand{\A}{\mathcal{A}}
\newcommand{\M}{\mathcal{M}}
\newcommand{\T}{\mathcal{T}}
\newcommand{\rew}{r}
\newcommand{\sinit}{s_t^{{\textsc{init}}}}
\newcommand{\sgoal}{s_t^{{\textsc{goal}}}}
\newcommand{\sfinal}{s_t^{{\textsc{final}}}}
\newcommand{\sprop}{s_t^{{\textsc{prop}}}}

\subsection{Problem Definition}
\label{subsec:problem_def}
We formulate the dexterous distributed manipulation problem as a Markov Decision Process (MDP) given by $\mathcal{M}=(\St, \A, \T, \rew, \gamma)$, where
$\St = \prod_{i} \St^i$ is the  state space and $\St^i$ represents the individual state space of robot $i$, and  $\A = \prod_{i} \A^i$ is the  action space and  each $\A^i$ represents the individual action space of robot $i \in \{1, 2, ... 64\}$. At each timestep $t$, we randomly generate an initial pose and a goal pose for an object being manipulated. We use an image processing pipeline to obtain pairs of robots and object-boundary points using nearest neighbor search, and eliminate robots directly on top of the objects and whose workspaces do not overlap with the object boundary as shown in Appendix \ref{app:vis_servo}. We define this neighborhood of robots with their indices sampled as a set $\N_t \subseteq \{1, 2, ... 64\}$. Each robot's state contains a set of initial boundary points $\sinit$, the position of the robot $\sprop$, and the set of goal boundary points $\sgoal$, resulting in a robot state space of $\St^i\in \mathbb{R}^{2+3+2}$. The actions taken by the robots are described as a set of 3D displacement vectors $a_t=(a_x,a_y,a_z) \in \mathbb{R}^3$. For planar manipulation tasks, we consider the 2D component of the action space $a_{t_\text{xy}}=(a_x, a_y)\in\mathbb{R}^2$, while $a_z\in \{\zlow,\zhigh\}$ is used as an action selection mechanism that determines $\Nsel\subseteq\N_t$ with $a_z=\zlow$. The dynamics of the environment are captured by the transition function $\T(s_{t+1}|s_t, a_t)\in\St\times\A\times\St\rightarrow[0,1]$, and the rewards are captured by the function $\rew(s_t, a_t)\in\St\times\A\rightarrow\mathbb{R}$. Finally, our objective is to learn policies that maximize the expected cumulative reward: $J(\pi) = \mathbb{E}_{s_t, a_t\sim}\left[\sum_{t=0}^{\infty} \gamma^t \rew(s_t, a_t)\right]$

\begin{figure*}[t!]
    \centering
    \includegraphics[width=0.99\textwidth]{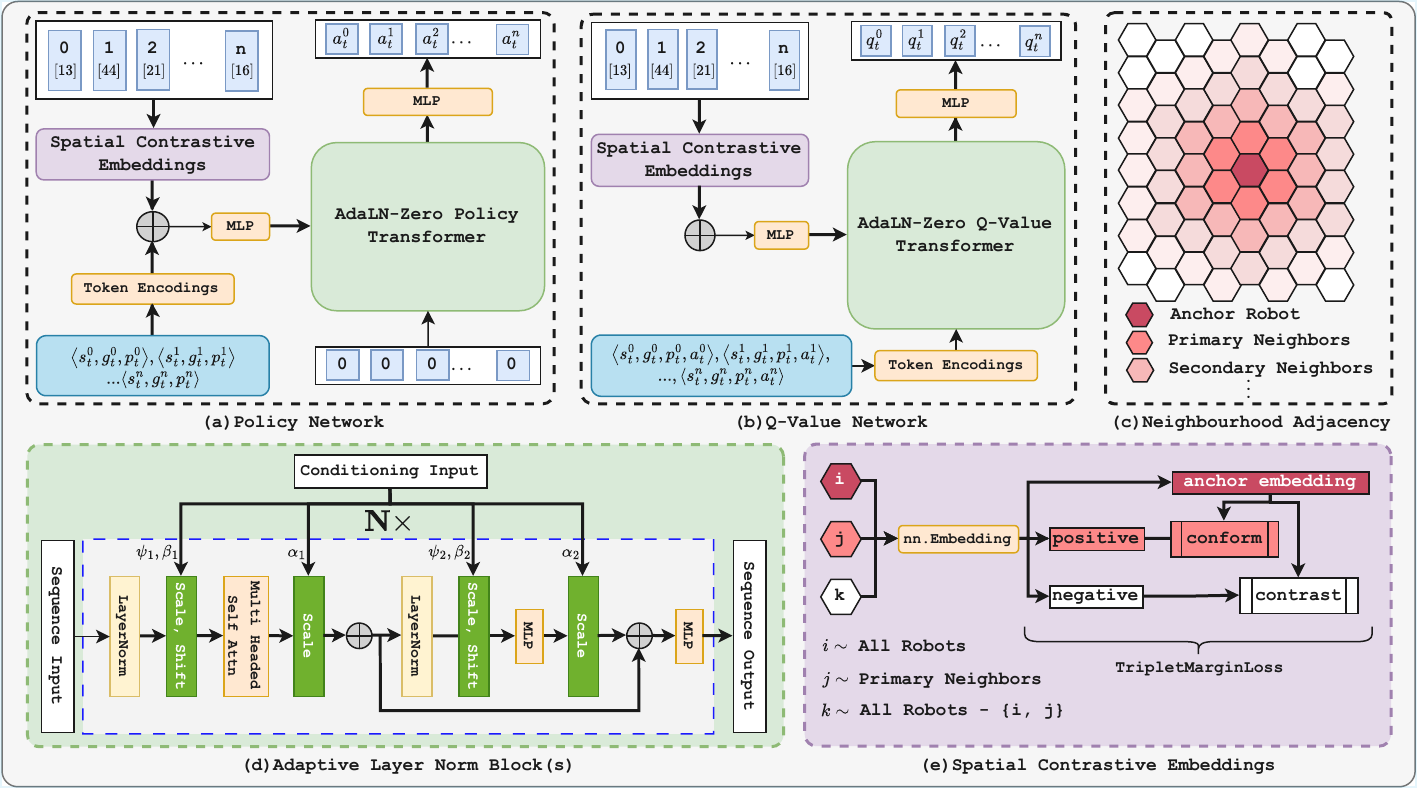}
    \caption{(a) Decoder-only MAT policy network. (b) Decoder-only MAT Q-value network. (c) Neighbourhood adjacency to select positive embeddings as primary neighbors and negative embeddings as farther away neighbors. (d) Adaptive Layer Norm Block. (e) Spatial Contrastive Embeddings using Triplet Margin Loss to create pairwise conformity and contrast in the embedding space.}
    \label{fig:method}
    \vspace{-0.5cm}
\end{figure*}

\section{Methods}
\label{sec:method}
\vspace{-0.4cm}

\subsection{Multi-Agent Transformers}
\label{subsec:mat_arch}
The core of our policy representation is a Multi-Agent Transformer (MAT) \cite{MAT}. Specifically, we use a \textit{decoder-only} transformer architecture as a policy and a Q function. The policy passes the zero vector of joint actions $a_t^{0_{0:n-1}}$ to a sequence of decoder transformer blocks. Actions are iteratively conditioned on the state and position embeddings through the attention layers of the transformer (Fig. \ref{fig:method}(a)). On the other hand, Q functions learn a joint distribution over the states and the actions to learn the expected reward distribution. Hence, we concatenate states and actions over all the robots as the sequential input to the transformer. These are conditioned on the position embeddings to obtain individual Q-values over all robots, mapping $\mathbb{E}[\rew]\gets \mathbb{E}[Q(s_t^i,a_t^i)]$ (Fig. \ref{fig:method}(b)). 


For efficient policy learning in high-dimensional \DDM tasks, we incorporate Adaptive Layer Normalization with Zero-Initialization (AdaLN-Zero) \cite{DiT, film, adaln_zero} as the transformer blocks. As shown in Fig.~\ref{fig:method}(d), each AdaLN-Zero layer takes the primary sequence (robot states/actions) and a conditional input to compute adaptive scaling ($\psi$) and shifting ($\beta$) parameters. Zero-initializing the learned scaling parameter $\psi$ accelerates training \cite{DiT}. A learned scaling parameter $\alpha$ is added before the residual connection for stability. $\psi$ and $\beta$ map conditional inputs to the sequential inputs, which mitigates the need to learn a complex attention map capturing representations with dynamically varying sequences of robots. 



During training, we use masked Multi-Head Attention (MHA) as proposed in \cite{maskedAttn} within the transformer blocks to allow agents to implicitly coordinate by attending to relevant information from others in the sequence. Crucially, the attention mechanism is guided by Spatial Contrastive Embeddings (SCEs), which ground the representations in the local neighborhood of the \deltaarrays as shown in the Appendix Fig. \ref{fig:pos_embeds}.
The final output of the MAT is a set of refined tokens, decoded into agent-specific actions $a_i \in \A$.

\begin{figure}[t!]
    \centering
    \includegraphics[width=0.99\textwidth]{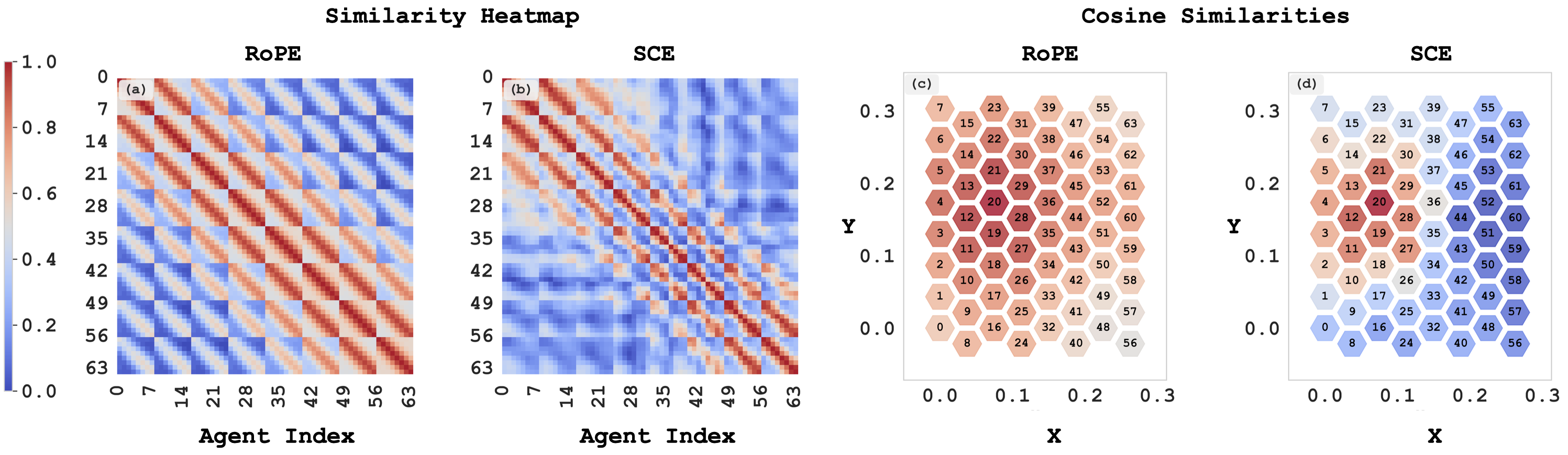}
    \caption{RoPE fails to generate embeddings that capture similarities among neighboring robots as SCE embeddings are able to do. (left) Heatmap of the cosine similarity among all the robots' embeddings. (right) Cosine similarity of an arbitrary robot (20) visualized on the delta array structure.}
    \label{fig:pos_embeds}
    \vspace{-0.5cm}  
\end{figure}

\subsection{Spatial Contrastive Embeddings}
\label{subsec:SCEs}
A fundamental challenge of \DDM systems is the stochastic ordering of the neighborhood for robots. At each timestep $t$ neighboring robots $\mathcal{N}_t$ are arbitrarily sequenced detached from their spatial arrangement in the real world. Standard position embeddings like Sinusoidal Position Embeddings (SPEs)\cite{vaswani} and Rotary Position Embeddings (RoPEs) \cite{RoPE} fail to capture spatial relationships in such dynamically changing agent configurations. We introduce Spatial Contrastive Embeddings (SCEs) to address this challenge. SCEs map each robot's index to a learned embedding vector $e_i \in \mathbb{R}^d$, creating a persistent spatial representation. 
These embeddings are trained offline using a triplet loss function:
\begin{equation}
\mathcal{L}_{\text{triplet}}(e_i, e_j, e_k) = \max(0, \|e_i - e_j\|_2^2 - \|e_i - e_k\|_2^2 + m)
\end{equation}
where $\text{e}_i$ and $\text{e}_j$ are embeddings of physically adjacent robots, $\text{e}_k$ are distant, and $m$ is a margin parameter. This ensures that neighboring robots have similar embeddings while distant robots have dissimilar ones, preserving the array's spatial topology in the learned representations. We compare the 2D representations projected by RoPE and SCE in Fig. {\ref{fig:pos_embeds}}. SCEs serve as conditional inputs to the AdaLN-Zero layers in the MAT, enabling the transformer to reason about spatial relationships through attention even when the active robot subsets vary dynamically

\subsection{Learning Phase 1: Pre-training via Behavior Cloning (\matbc)}
\label{subsec:mabc}
To improve sample efficiency and provide a good initialization for policy learning in the highly redundant \deltaarrays action space, we first pre-train the MAT policy using Behavior Cloning (\matbc). However, providing human demonstrations for a 64-robot system is non-trivial due to the complexity of teleoperating numerous agents simultaneously. Hence, we develop a visual servoing pipeline to generate \emph{expert} demonstrations. Following the visual servoing pipeline described in Appendix Sec. \ref{app:vis_servo} to generate a dataset of \emph{expert} demonstrations $\rb_{exp}$
The \matbc policy, $\pi_\phi^{BC}$, is trained by minimizing the Mean Squared Error (MSE) between its predicted actions $\A^\pi = \pi_\phi^{BC}(\St)$ and the expert actions $\A^e$.
Concurrently, we pre-train an initial critic network, $Q_\theta^{BC}$. This critic is trained via Mean Squared Error (MSE) on a mixed dataset $\rb_{mixed}$, containing both expert and random transitions,  to predict the immediate reward $r$ associated with a state-action pair. During BC, our Q function does not incorporate future rewards via Bellman updates due to the quasi-static nature of visual servoing data. This provides a basic value estimation before RL fine-tuning. The specific loss function is detailed in Appendix Sec. ~\ref{app:losses}.

\subsection{Learning Phase 2: Fine-tuning via Reinforcement Learning (\matbcft)}
\label{subsec:matsac}
In the second phase, we train the policy and the critic using Soft Actor-Critic (SAC) \cite{SAC} adapted for multi-agent transformers (\matsac) as a baseline. We initialize the actor and critic networks with the pre-trained weights: $\pi_\phi \leftarrow \pi_\phi^{BC}$ and $Q_\theta \leftarrow Q_\theta^{BC}$. We refer to the resulting fine-tuned policy as \matbcft. The critic $Q_\theta(s_t, a_t)$ takes the global state $\St$ and joint action $\A$ and outputs agent-specific Q-values $[q_1, \dots, q_N]$. To avoid overestimation bias, we train two critics ($Q_{\theta_1}, Q_{\theta_2}$) and use the minimum for target calculations. The actor $\pi_\phi$ is trained to maximize both expected future returns and entropy, facilitating exploration in the high-dimensional action space. 

To address execution costs of controlling the robots, we shape reward functions with a discrete and a continuous cost over the policy's action selection. Let $\Nsel \subseteq \N$ be the selected robots with $a_z=10cm$, and $\asel$ be their corresponding 2D actions. We study four rewards to select a sparse set of robots to execute movements for each step we term as \textbf{action selection}: 
\begin{enumerate}
    \item The original reward function (OG): $\rog \leftarrow \frac{1}{c \delta^2 + \epsilon}$, where $\delta$ is given by $||\sgoal-\sfinal||_2^2$, $c$ is a scaling factor that controls the shape of the reward function, and $\epsilon$ is a bounding factor. We set $\epsilon=0.01$ to bound the rewards between $0$ and $100$
    
    \item Discrete Execution Cost (DEC): $\rdec \leftarrow \rog - \lambda_1 \frac{|\Nsel|}{|\N|}$
    \item Continuous Execution Cost (CEC): $\rcec \leftarrow \rog - \lambda_2 \sum_{i \in \Nsel} ||\asel||_2$
    \item Merged Execution Cost (MEC): $\rgec \leftarrow \rog - \lambda_1 \frac{|\Nsel|}{|\N|} - \lambda_2 \sum_{i \in \Nsel} ||\asel||_2$
\end{enumerate}

Where $\lambda_1$ and $\lambda_2$ are weighting hyperparameters that control the trade-off between task performance and resource efficiency. The policy learns that robots with $a_z = \zhigh$ do not interact with the objects and can mitigate unnecessary collisions due to crowding. The complete loss functions and SAC algorithm details are provided in Appendix Sec.~\ref{app:matsac_losses}.

\section{Experiments}
\label{sec:expts}
We train all our policies in simulation and evaluate our proposed methods zero-shot for \DDM tasks using the delta arrays (Sec.~\ref{sec:bgnd}) in both MuJoCo and on the real hardware. Our evaluation follows a set of pre-defined trajectories consisting of 20 subgoals. The task is closed-loop SE(2) manipulation of 10 different objects (shown in Appendix~\ref{app:gt_trajs}) from an initial pose to a subgoal pose sequentially. We measure success rate (reaching a subgoal within 3 attempts) and mean trajectory tracking error ($\mu \pm \sigma$) using quasi-static closed-loop rollouts. Aggregate results are presented in Sec.~\ref{sec:results}, with object-specific details in Appendix~\ref{app:obj_perf}. Specifically, we aim to answer 3 questions through our experiments:

\textbf{Q1:} \textit{What is the significance of position embeddings for learning \DDM policies?}

\vspace{-0.1cm}
Transformers typically rely on position embeddings to incorporate sequential structures. Sinusoidal Position Embeddings (SPEs) use sine and cosine functions of frequencies along a sequence, which is beneficial for natural language tasks. While Rotary Position Embeddings (RoPE) mix pairs of coordinates in an outward-growing helical pattern, which is beneficial for fixed grid inputs. SCEs and RoPE are at the two ends of a spectrum from pretrained embeddings to scalable transformations, respectively. Hence, we ablate three intermediate position embeddings: (1) A locally constrained RoPE (\lcrope) which zeroes out the attention activations beyond a distance threshold, (2) A learned relative embedding (\lre) that maps $K$ neighbors of a robot to a fixed set of learnable angle vectors, and constructs a relative rotation matrix consistent across different robots, and (3) Use of integer position embeddings just like SCE, but without pretraining (\learned). We follow the \matbcft pipeline to pretrain and finetune the policies with all the position embedding ablations. 

\textbf{Q2:} \textit{How does the selection of attention mechanisms affect the wall-clock speed and sample-efficiency of learning \DDM policies?}

\vspace{-0.1cm}
Standard transformer architectures use self-attention \cite{vaswani} and cross-attention \cite{crossattn} blocks stacked on top of each other to learn shared representations between sequential and conditional inputs. However, recent works like \citet{adaln_diff, DiT} have demonstrated the effectiveness of Adaptive Layer Normalization (AdaLN) to speed up the training of large autoregressive diffusion transformers. We compare AdaLN-Zero against standard self-attention and cross-attention in terms of training stability, speed of convergence using \matsac.

\textbf{Q3:} \textit{What are the tradeoffs of inducing policies to use fewer robots on the ability of the \deltaarrays to perform long-horizon closed-loop planar manipulation tasks?}

\vspace{-0.1cm}
While achieving high task performance is paramount, the efficiency of execution is also critical for \DDM systems, especially in the real world. Effectively selecting a subset of robots ($\Nsel$) for a given manipulation step can reduce energy consumption, computational load during inference, physical wear, and the risk of inter-robot collisions. To study the effect of the action selection rewards in \ref{subsec:matsac}, we train four \matbcft policies: \matbcftog, \matbcftdec, \matbcftcec, \matbcftgec. We compare these methods with visual servoing, \matbc, \matsac trained from scratch, and \matbcft with all robots selected, and quantify the tradeoff in terms of the average number of active robots selected, tracking error along the trajectory, and success rate in achieving subgoals along the trajectory. Implementation details, hardware specifications, and training protocols are in the Appendix Sec. ~\ref{app:impl}.

\section{Results and Discussion}
\label{sec:results}

        
        

\begin{figure}[t!]
    \centering
    \includegraphics[width=\textwidth, height=0.25\textheight, keepaspectratio]{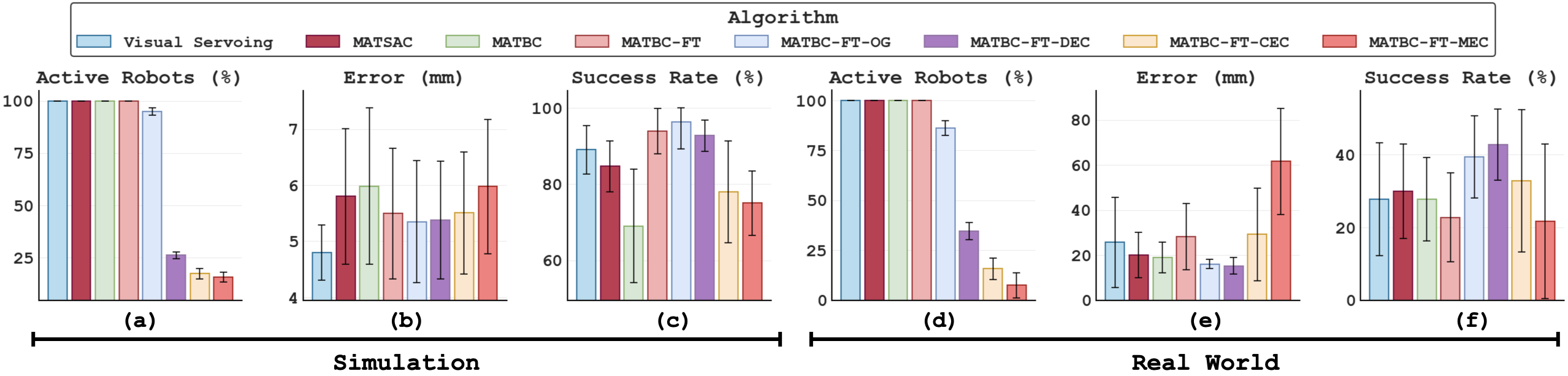}
        
    \caption{Overall performance for three rigid objects (hexagon, star, trapezium) on (a - c) Simulation, (d - f) Real World. For each, (Left) shows the percentage of active robots involved in each manipulation step. (Middle) shows the error at every execution step. (Right) shows the percentage success rate of reaching subgoals in 3 tries over the trajectories}
    \label{fig:overall}
    \vspace{-0.5cm}
\end{figure}

\begin{figure}[h!]
    \centering
    \includegraphics[width=\textwidth, height=0.25\textheight, keepaspectratio]{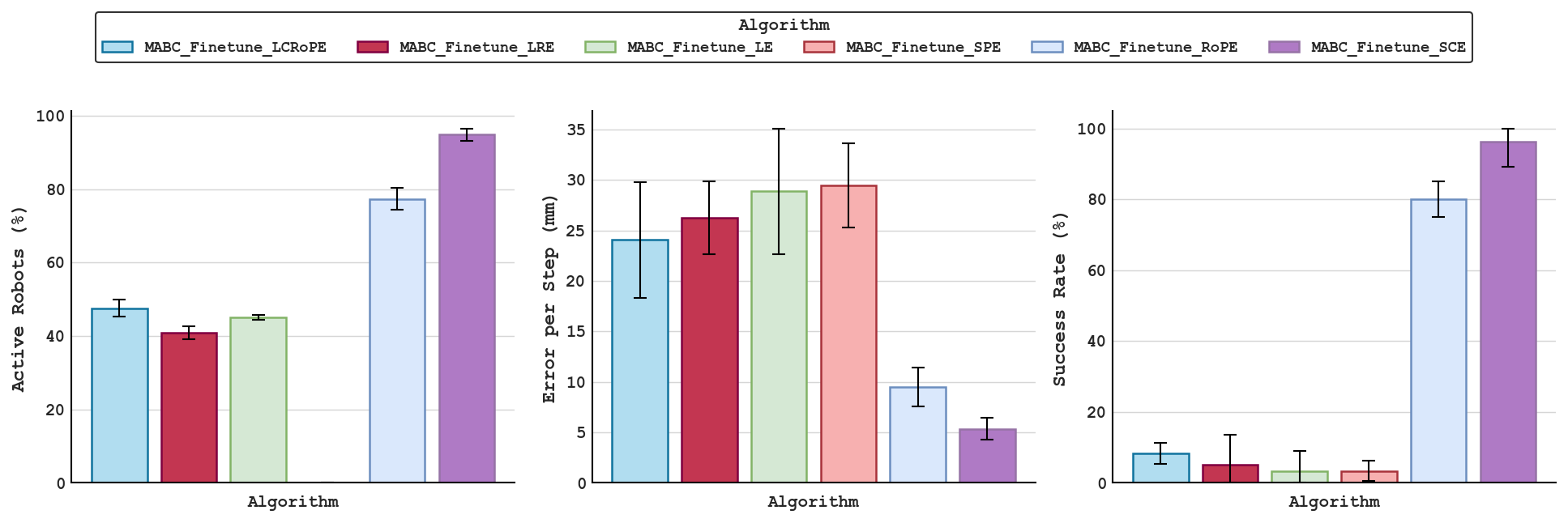}
        
    \caption{}
    \label{fig:pos_embed_ablations}
\end{figure}

\textbf{6.1 Significance of Spatial Embeddings:} 
Our experiments with training SPE and RoPE fail to converge while training \matsac from scratch. Hence, our key ablations are performed on the \matbcftog pipeline. Using Fig. \ref{fig:pos_embed_ablations} as reference, we see that SPE still fails to learn any meaningful policies. However, RoPE shows dramatic improvement over RL training from scratch when the initial exploration is mitigated. Our ablations with \lcrope, \lre, and \learned resulted in poorly trained policies that fail to accomplish the trajectory object pushing task. The key takeaways are: (1) Learning position embeddings online is hard due to poor multimodal reasoning capabilities of AdaLN-Transformers. (2) Structured relative spatial bias over the entire attention mechanism is crucial for learning centralized MARL policies for \DDM tasks. (3) Pretraining SCEs provides a highly contrastive conditioning to the attention mechanism, compared to the relatively similar weighing functions provided by RoPE. 

\begin{figure}[h!]
    \centering
    \subfloat[]{\includegraphics[width=0.48\textwidth]{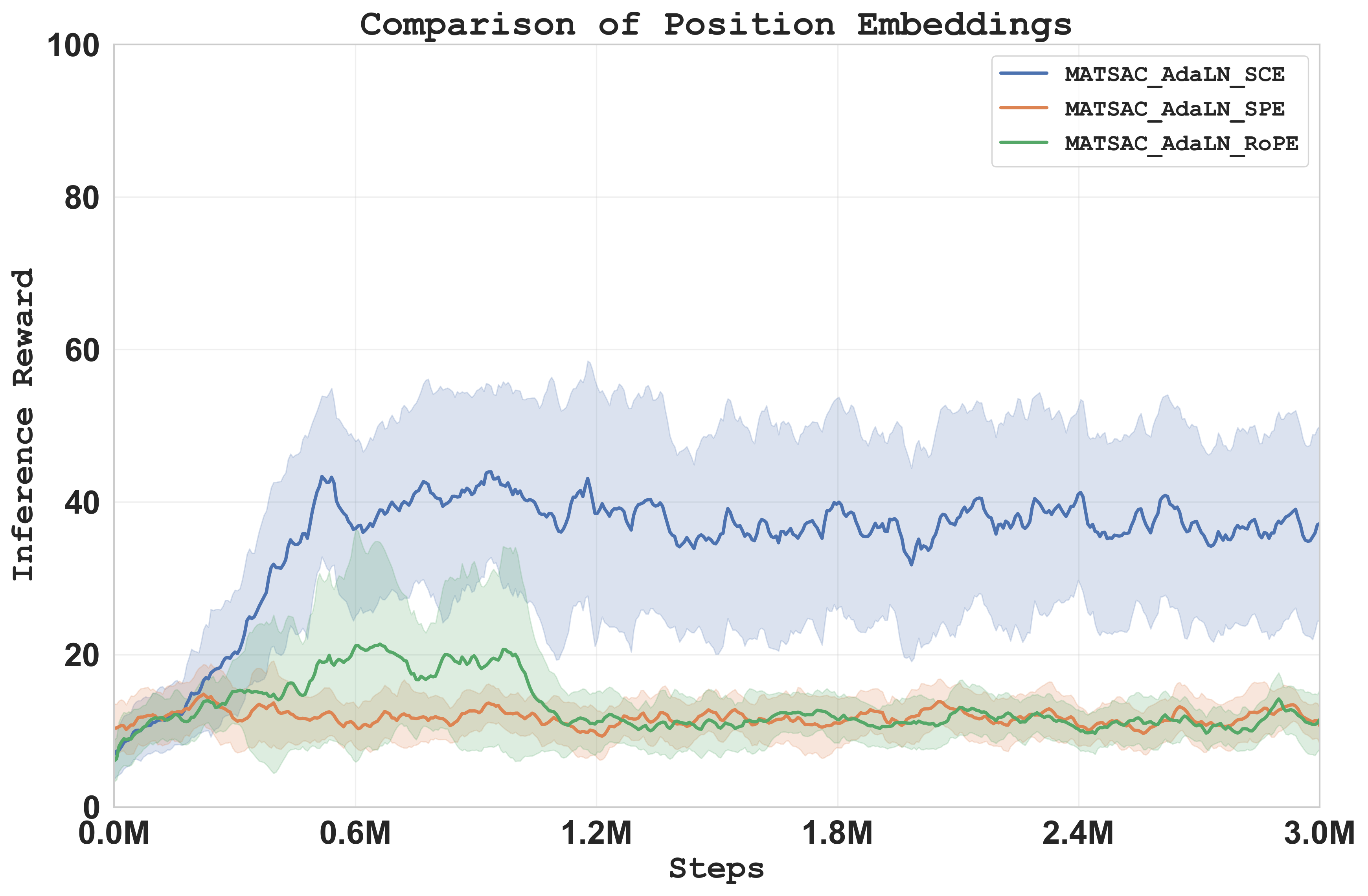}
    \label{fig:comp_pos_embed}}
    \hfill
    \subfloat[]{\includegraphics[width=0.48\textwidth]{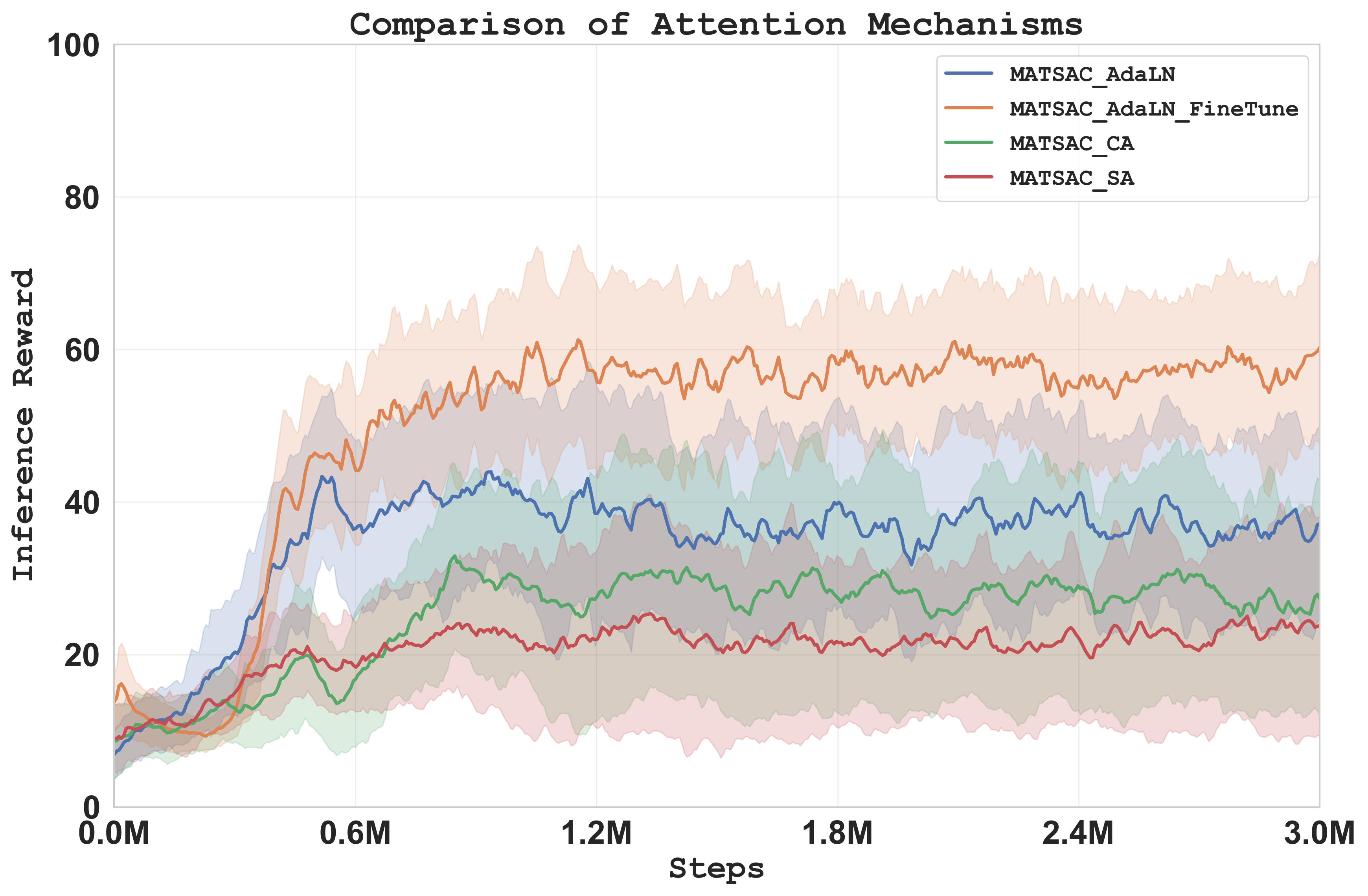}
    \label{fig:comp_attn}}
    \caption{\textbf{Inference Reward Plots} a) Attention Mechanisms - AdaLN performs better than Cross Attention and Self Attention. AdaLN transformer fine-tuned using SAC performs better than training tabula-rasa. b) Position Embeddings - Sinusoidal and Rotary Embeddings fail to learn spatial correlations among neighboring sets of robots while SCEs demonstrably do so.}
    \label{fig:training_plots_comparison}
\end{figure}

However, they do not adequately inform the spatial arrangement of robots to condition the actions, especially when a subset of them are provided as input in arbitrary order. Unlike language and a fixed grid over images, robot selection in multi-agent settings is stochastic. We hypothesize that using predetermined embeddings to represent a dynamically varying order of the sequential input induces multi-modality in the learning pipeline, which makes it hard for policy gradient methods to converge \cite{multimodality_pg2, multimodality_pg}. As such, policies trained using Sinusoidal Position Embeddings (SPEs) or Rotary Position Embeddings (RoPEs) failed to converge and achieve meaningful task performance (Fig.~\ref{fig:comp_pos_embed}). 



\textbf{6.2 Efficiency of Attention Mechanisms:}
\label{subsec:results_q2}
Our results show that using AdaLN-Zero layers leads to significantly faster and more stable training compared to standard Self-Attention (SA) or Cross-Attention (CA) blocks (Fig.~\ref{fig:comp_attn}). The wall-clock training time for AdaLN-Zero was approximately half that of SA/CA ($\sim$23 vs. $\sim$44 hours), consistent with findings in other domains \cite{DiT, bczero}. Furthermore, AdaLN-Zero achieved better final policy performance (higher average reward). This validates our choice of using AdaLN-Zero for computational efficiency without sacrificing performance in highly redundant \DDM tasks.

\textbf{6.3 Trajectory Tracking Performance and Action Selection Tradeoff:}
We evaluate the core learning pipeline and the impact of action selection strategies in both simulation and through zero-shot transfer to real hardware (Fig.~\ref{fig:overall}). We define success as the object reaching the goal within a mean Euclidean distance of 7.5 mm over the 2D boundary points of the object. Since visual servoing actions are 2D vectors, the training data for behavior cloning are 2D vectors as well. \matbc, \matsac, and \matbcft engage all robots in $\N_t \leq 64$. MATBC pretrains on the visual servoing data with $a_z$ always to $\zlow$ to obtain a pretrained policy. During finetuning, our action selection mechanism learns to select $\Nsel \subseteq \N_t$. Only the robots with $a_z = \zlow$ manipulate the object, and $a_z = \zhigh$ disengage. This reduces the active number of robots from $|\N_t|$ to $|\Nsel|$. \textit{MATBC-FT-}(\textit{OG, DEC, CEC, MEC}) all allow learned $a_z$ values, and thus trade off between the number of robots selected and the task performance.

\begin{figure}[t!]
    \centering
    \includegraphics[width=0.97\textwidth]{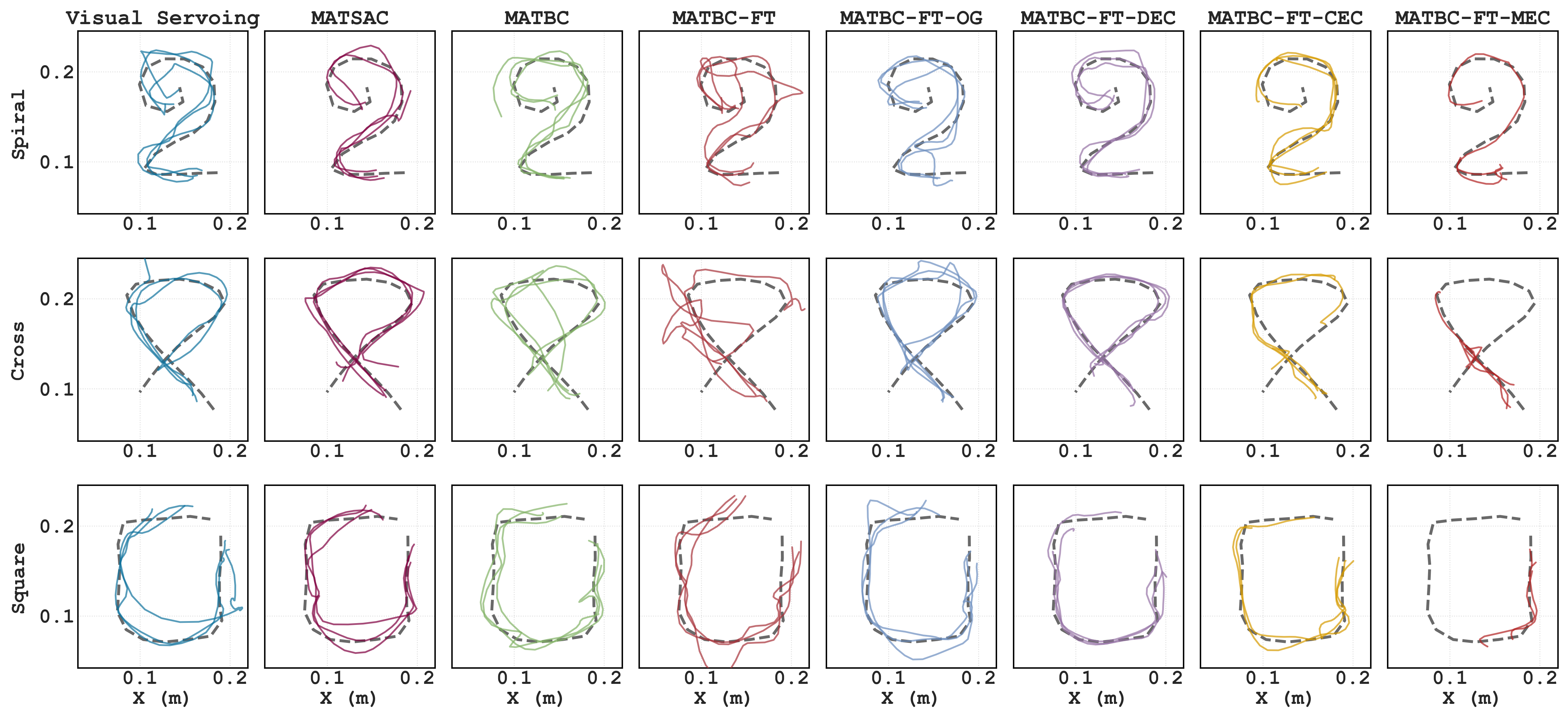}
    \caption{Trajectory tracking results on our three hardest trajectories with three objects -- hexagon, trapezium, star -- in the real world.}
    \label{fig:test_traj_track_real}
    \vspace{-0.5cm}
\end{figure}

RL fine-tuning (\matbcft) consistently improves upon \matbc alone, and \matsac from scratch, which is consistent with robot learning literature \cite{real1, Xu2022}. \matbcft achieves the lowest tracking error and highest success rate among methods using full $\N_t$. However, when action selection is enabled, \matbcftog marginally improves upon \matbcft in sim ($\sim1.8\%$), but significantly improves in the real world ($\sim40.7\%$) in average tracking error. We attribute this to deterministic physics in sim causing the performance improvement to saturate. Moreover, comparing \matbcftog with \matbcftdec, \matbcftcec, and \matbcftgec, we conclude that adding a discrete penalty to the reward formulation encourages MATs to better leverage the tradeoff between execution cost and planar manipulation performance. Fig. \ref{fig:test_traj_track_real} shows that \matbcftcec and \matbcftgec really struggle with closed-loop long-horizon \DDM tasks. On the other hand, \matbcftdec faithfully tracks objects along the desired trajectory, while using $\sim55\%$ fewer robots compared to \matbcftog, and $\sim65\%$ fewer than \matbcft. 

\begin{figure}[h!]
    \centering
    \includegraphics[width=\textwidth, height=0.25\textheight, keepaspectratio]{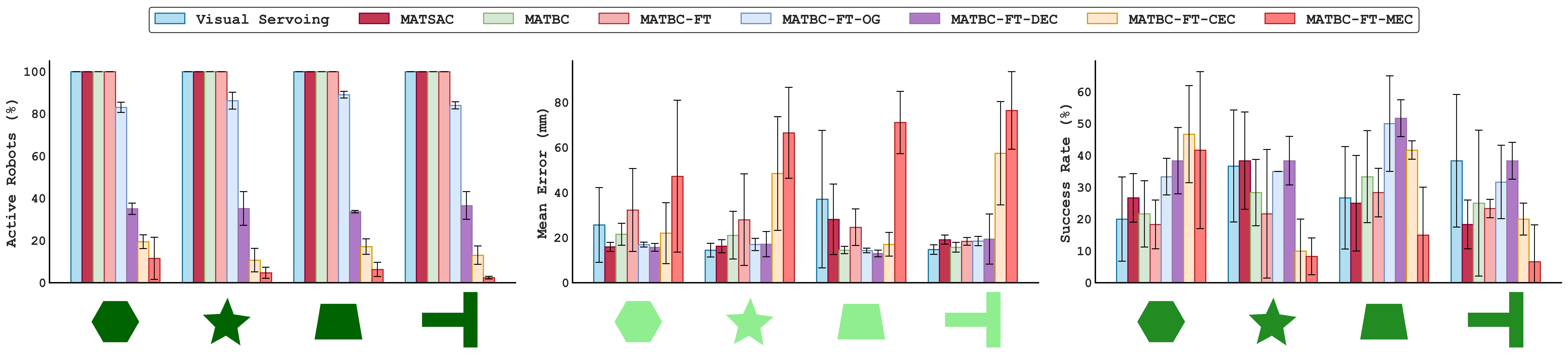}
        
    \caption{Object level comparison between all the methods in the real world. The T-shaped object was unseen during training. (a) Shows percentage of active robots involved in each manipulation step. (b) Shows the error at every execution step. (c) Shows the number of attempts needed by an algorithm to complete a 20-step trajectory. }
    \label{fig:real_world_with_unseen}
\end{figure}

\textbf{6.4 Qualitative Evaluation:} We observe significantly high policy performance in deterministic settings, as are simulation environments, compared to the stochasticity of soft robot kinematics in the real world. The soft delta robots in the \deltaarrays demonstrate a spring-damper behaviour when multiple robots manipulate a single object. This makes execution in MuJoCo highly stable, but creates practical challenges in the real world due to the making and breaking of compliant contacts induce out-of-distribution errors during trajectory tracking for the objects. Hence we see a significant drop in success rates of \matbcft and \matbcftog, which use a higher number of robots, compared to the relatively minor drop of \matbcftdec. This can also be qualitatively seen in Fig. \ref{fig:test_traj_track_sim} and Fig. \ref{fig:test_traj_track_real}. We further show out-of-distribution object generalization by deploying the full suite of methods on a push-T task in Fig. \ref{fig:real_world_with_unseen}

\section{Conclusion}
\label{sec:conclusion}
In this work, we presented an initial exploration toward learning robust and sample-efficient policies for Distributed Dexterous Manipulation (\DDM) on \deltaarrays using spatially conditioned MAT. We demonstrate how design choices like attention mechanisms and position embeddings significantly affect the performance of training multi-agent SAC using MATs. Further, we demonstrate that incorporating an action selection mechanism with a discrete penalty (\matbcftdec) enables the policy to learn an effective policy that maximizes performance and efficiency compared to continuous penalties. \matbcftdec reduces active robot usage by up to 65\% while maintaining $\sim1.5$ cm average trajectory tracking error with a soft robot array. Moreover, action selection significantly mitigates robotic wear and tear in real-world deployment.



\clearpage

\section{Limitations}
\label{sec:limits}
In this paper, we proposed multi-agent policy learning for DDM as a novel problem formulation. Hence, we limited the difficulty of tasks to 2D planar manipulation. Studying the efficiency of these methods to learn DDM policies in 3D space remains an unexplored problem. On the other hand, although fine-tuning shows a major improvement in sample efficiency, it still needs expert demonstrations to pretrain the behavior cloning policy, which is challenging for the multi-robot cooperation domain. Expert demonstrations will be harder to obtain in 3D space. We identify tabula rasa training not being able to match the performance of fine-tuned methods as the second limitation. Moreover, we can show our experiments only on the delta arrays, as the soft robots allow for such experimental setups to be made possible. This can be considered a limitation in terms of the demonstration of the generalizability of the proposed method for \DDM tasks. Diffusion models have been extensively used in behavior cloning, but RL-based fine-tuning of multi-agent diffusion policies remains an unexplored avenue, especially for highly redundant action spaces as those of the delta arrays. Finally, in this paper, we explored the use of multi-agent transformers for \DDM tasks, which can be thought of as a centralized method for policy learning. Understanding the fundamental differences in decentralized vs centralized methods for distributed dexterous manipulation can a fundamental area of future work. 

\clearpage



\bibliography{references}  

\clearpage

\section{Appendix}
\subsection{Visual Servoing}
\label{app:vis_servo}
Providing human demonstrations for a 64-robot system is non-trivial due to the complexity of coordinating many agents. Hence, we propose a visual servoing pipeline to collect expert demonstrations $\rb_{exp}$ for \matbc pre-training (Sec.~\ref{subsec:mabc}).

First, we segment objects in the camera view using Language Segment Anything Model (Lang-SAM) \cite{SAM, groundingdino} for complex shapes or standard HSV filtering for simpler geometries. This yields a set of $M$ 2D boundary points $\B = \{b_1, b_2, \ldots, b_M\} \in \mathbb{R}^{M\times2}$. We then run a  nearest neighbor search algorithm to identify the set of robots $\N$ whose workspaces overlap with the object's boundary but whose base positions are outside the object's initial polygon. For each robot $i \in \N$, we identify its closest boundary point $b_i \in \B$.

For a given target pose (defined by a relative 2D translation and rotation), we compute the corresponding 2D rigid transformation matrix $T$. We compute the goal boundary points $\B'$ by applying this transformation to the initial boundary points: $\B' = T(\B) = \{T(b_1), T(b_2), \ldots, T(b_M)\}$. The corresponding goal point for robot $i$'s closest initial point $b_i$ is $b'_i = T(b_i)$.

The raw visual servoing action $a_i^e$ for robot $i$ is generated as the displacement vector (flow vector) from its corresponding initial boundary point $b_i$ to the goal boundary point $b'_i$:
\begin{equation}
    a_i^e = \text{clip}(b'_i - b_i, a_{\text{min}}, a_{\text{max}})
\end{equation}
where $a_{\text{min}}, a_{\text{max}}$ are the workspace limits of the delta robots.

\subsection{Additional Loss Function Details}
\label{app:losses}


\subsubsection{Behavior Cloning (BC) Loss Functions}
The \matbc policy, $\pi_\phi^{BC}$, is trained by minimizing the Mean Squared Error (MSE) between the predicted actions $a_{t,i}^\pi = \pi_\phi^{BC}(s_t)_i$ and the expert actions $a_{t,i}^e$ for each agent $i$ in the neighborhood $\N_t$:
\begin{equation}
\label{eq:bc_appendix}
J_{BC}(\phi) = \mathop{\mathbb{E}}\limits_{(s_t, a_t^e)\sim\rb_{exp}} \left[ \sum_{i \in \N_t} \frac{1}{2} || a_{t,i}^\pi - a_{t,i}^e ||_2^2 \right]
\end{equation}
Note: $a_t^e = [a_{t,1}^e, \dots, a_{t,N_t}^e]$ is the joint expert action. $a_{t,i}^\pi$ is the action predicted for agent $i$ by the policy given the joint state $s_t$.

The initial critic network $Q_\theta^{BC}$ used for BC pre-training (Sec.~\ref{subsec:mabc}) is trained to predict the immediate global reward $r_t = \rew(s_t, a_t)$ using mean squared error loss on the mixed dataset $\rb_{mixed}$. Assuming a centralized critic predicting the global reward:
\begin{equation}
\label{eq:mabc_q_appendix}
J_{Q^{BC}}(\theta) = \mathop{\mathbb{E}}\limits_{(s_t, a_t, r_t)\sim\rb_{mixed}} \left[ \frac{1}{2} \left( Q_\theta^{BC}(s_t, a_t) - r_t \right)^2 \right]
\end{equation}
Note: $Q_\theta^{BC}(s_t, a_t)$ represents the predicted immediate global reward given the joint state $s_t$ and joint action $a_t$. (If the critic is intended to output per-agent values summing to $r_t$, the equation would need adjustment).

\subsubsection{Multi-Agent Soft Actor Critic Loss Functions}
\label{app:matsac_losses}
Assuming the critic decomposes the global value into per-agent Q-values $Q_{\theta_k}(s_t, a_t)_i$ for $i \in \N_t$, the critic loss for each of the two critics ($k=1,2$) is defined as:
\begin{equation}
\label{eq:matsac_q_appendix}
J_Q(\theta_k) = \mathop{\mathbb{E}}\limits_{(s_t, a_t, r_t, s_{t+1})\sim \rb} \left[ \sum_{i \in \N_t} \frac{1}{2} \left( Q_{\theta_k}(s_t, a_t)_i - \hat{q}_{t,i} \right)^2 \right] \quad \text{for } k=1,2
\end{equation}
where $\rb$ is the replay buffer containing environment interaction data, and the target Q-value $\hat{q}_{t,i}$ for agent $i$ at time $t$ is:
\begin{equation}
\label{eq:matsac_q_target_appendix}
\hat{q}_{t,i} = r_t + \gamma \mathop{\mathbb{E}}\limits_{a_{t+1}\sim\pi_\phi(\cdot|s_{t+1})}\left[ \min_{k=1,2} Q_{\bar{\theta}_k}(s_{t+1}, a_{t+1})_i - \alpha_{ent} \log \pi_\phi(a_{t+1,i}|s_{t+1})\right]
\end{equation}
Here, $\bar{\theta}_k$ are the target network parameters (updated via Exponential Moving Average (EMA)), $\gamma$ is the discount factor, and $\alpha_{ent}$ is the entropy temperature. $a_{t+1,i}$ is the action for agent $i$ in the next joint action $a_{t+1}$.

The actor loss is defined as:
\begin{equation}
\label{eq:matsac_pi_appendix}
J_\pi(\phi)=\mathop{\mathbb{E}}\limits_{s_t\sim \rb, \tilde{a}_t\sim\pi_\phi(\cdot|s_t)} \left[ \sum_{i \in \N_t} \left( \alpha_{ent} \log \pi_\phi(\tilde{a}_{t,i}|s_t) - \min_{k=1,2}Q_{\theta_k}(s_t, \tilde{a}_t)_i \right) \right]
\end{equation}
where $\tilde{a}_t = [\tilde{a}_{t,1}, \dots, \tilde{a}_{t,N_t}]$ are actions sampled from the Squashed Gaussian policy \cite{SAC} using the reparameterization trick for differentiability. $\pi_\phi(\tilde{a}_{t,i}|s_t)$ represents the probability of agent $i$'s action under the policy given the joint state $s_t$.

\subsection{Training Protocol and Hardware Specifications}
\label{app:impl}
We use 10-layer MATs for the policy and two Q-values, respectively, with 128-dim weights for all hidden layers. 
For MABC, visual servoing, as \emph{expert demonstrations}, is collected for 500 episodes per object in sim, and the pretraining phase runs for 400 epochs to obtain \matbc. 
All \matsac policies (including fine-tuning and training from scratch) are trained for 3,000,000 environment steps with a batch size of 256 and standard learning rates of 3e-4, discount factor $\gamma=0.99$. We use an Nvidia 4090 with an AMD EPYC 9554 CPU for all our experiments.

\clearpage
\subsection{Ablation Results}
\label{app:abl_results}
The following plots show the learning curves while training \matsac from scratch.

The following plots show the performance of \matbcftog with all the position embedding ablations. 
\subsection{Position Embedding Ablations}
\clearpage

\subsection{Experiment Trajectories}
\label{app:gt_trajs}
Ground truth trajectories which we manipulate the objects over:
\begin{figure}[h]
    \centering
    \includegraphics[width=0.99\textwidth, height=0.25\textheight,]{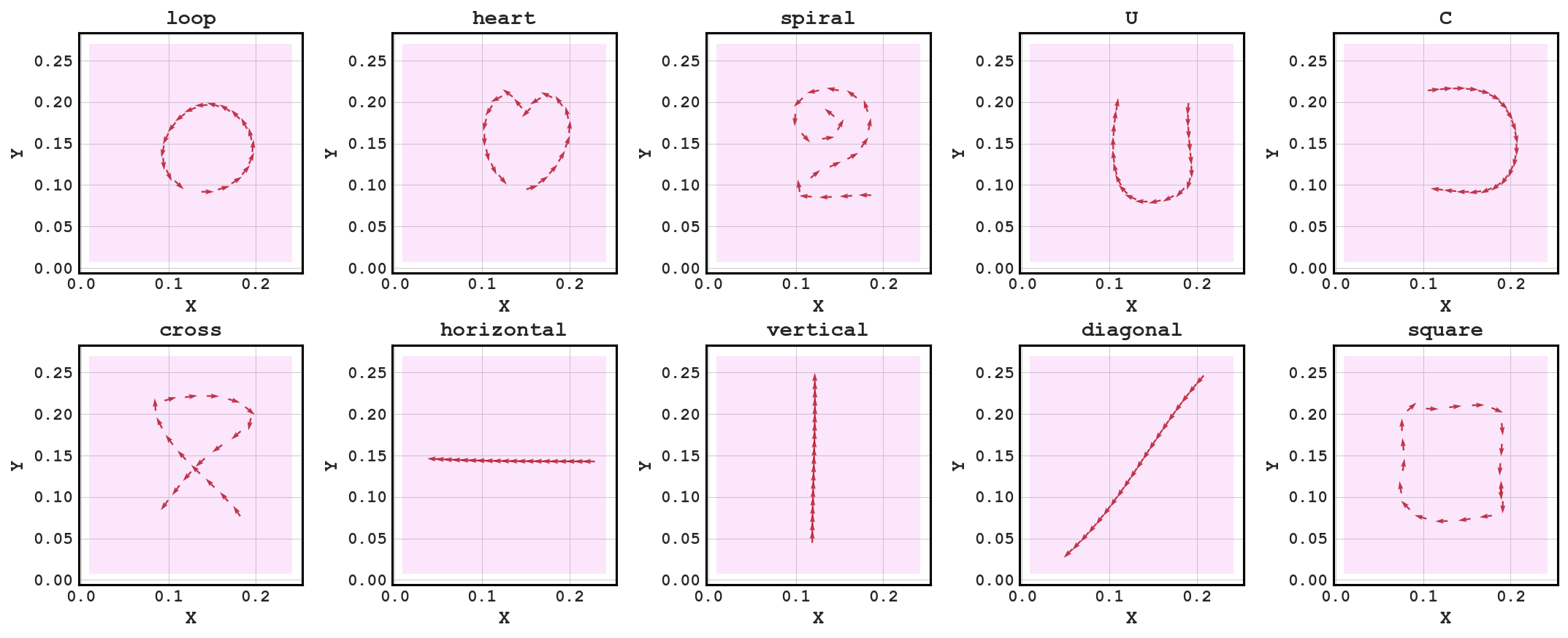}
    \caption{Ground truth trajectories defined as the inference task for closed-loop \DDM tasks.}
    \label{fig:test_traj_track_sim}
\end{figure}

\subsection{Object Level Performance - Real}

\clearpage

\subsection{Object Level Performance - Sim}
\label{app:obj_perf}
We show the performance of all algorithms on all objects averaged over all trajectories in simulation. The trends show that generally, convex objects are easier to manipulate by the delta arrays due to their workspace limitations, but they struggle with non-convex objects, especially during larger rotation angles. 

\begin{figure}[h!]
    \centering
    \subfloat[]
        {\includegraphics[width=\textwidth, height=0.25\textheight, keepaspectratio]{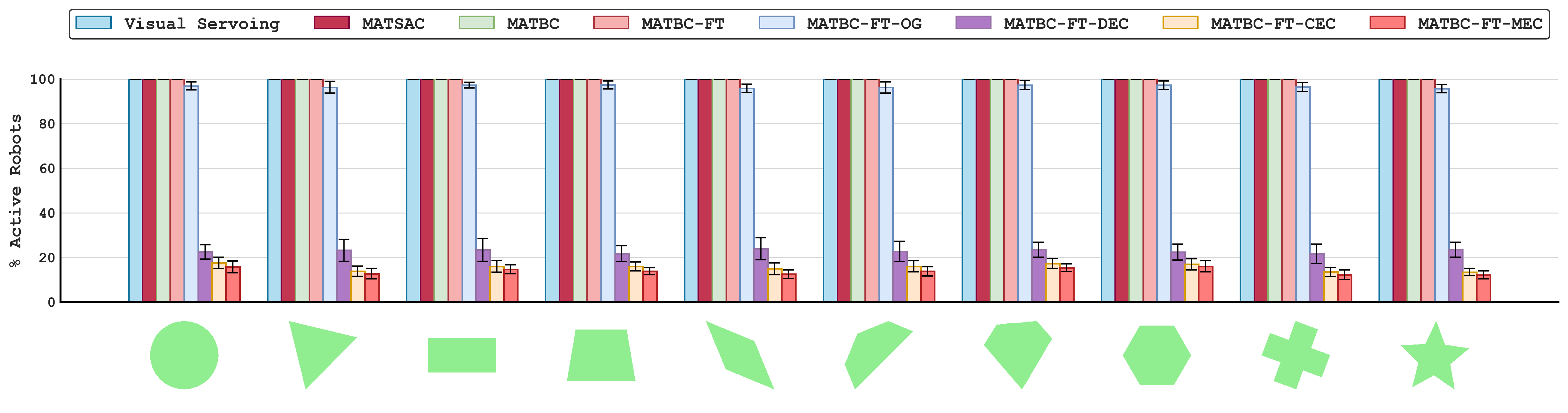}}
        
    \subfloat[]
        {\includegraphics[width=\textwidth, height=0.25\textheight, keepaspectratio]{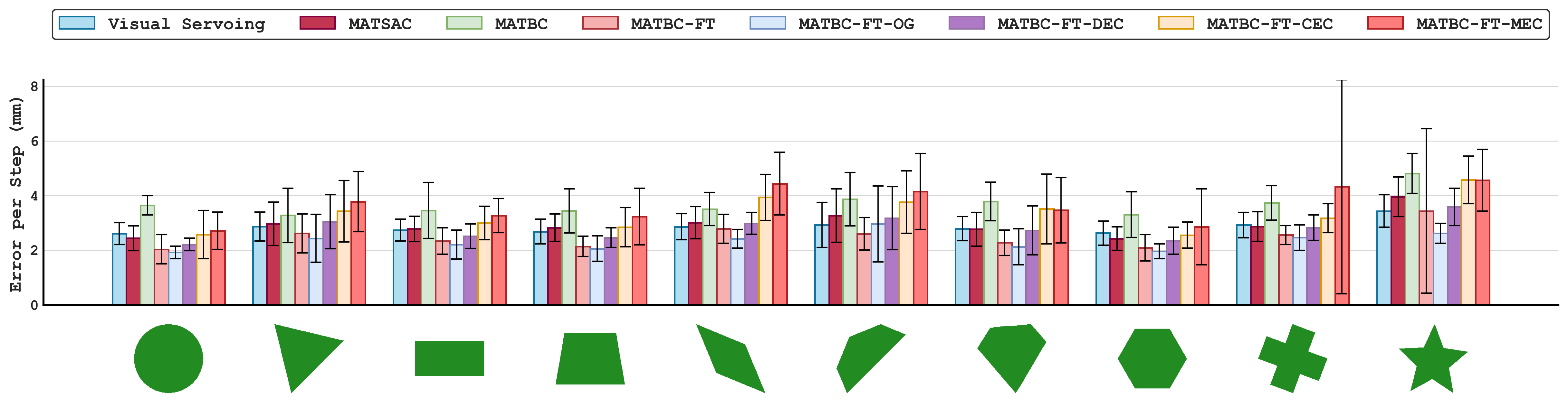}}
        
    \subfloat[]
        {\includegraphics[width=\textwidth, height=0.25\textheight, keepaspectratio]{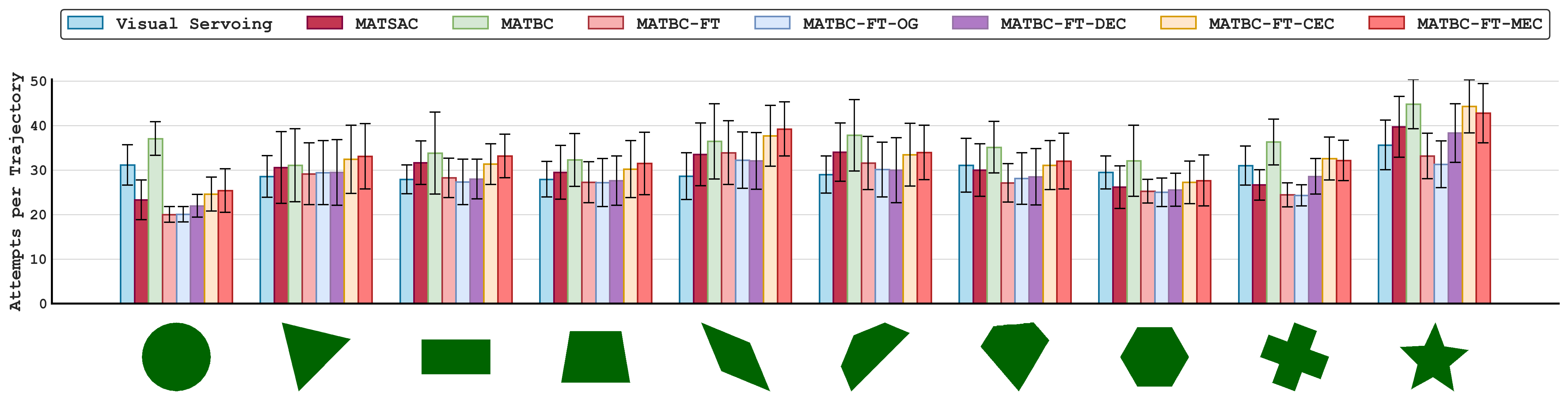}}
    \caption{Object level comparison between all the methods. (a) shows percentage of active robots involved in each manipulation step. (b) shows the error at every execution step. (c) shows the number of attempts needed by an algorithm to complete a 20-step trajectory. }
    \label{fig:obj_method_comparison}
\end{figure}

\clearpage

\subsection{Results Per Trajectory}
\label{app:traj_results}
\begin{figure}[h]
    \centering
    \includegraphics[width=0.99\textwidth]{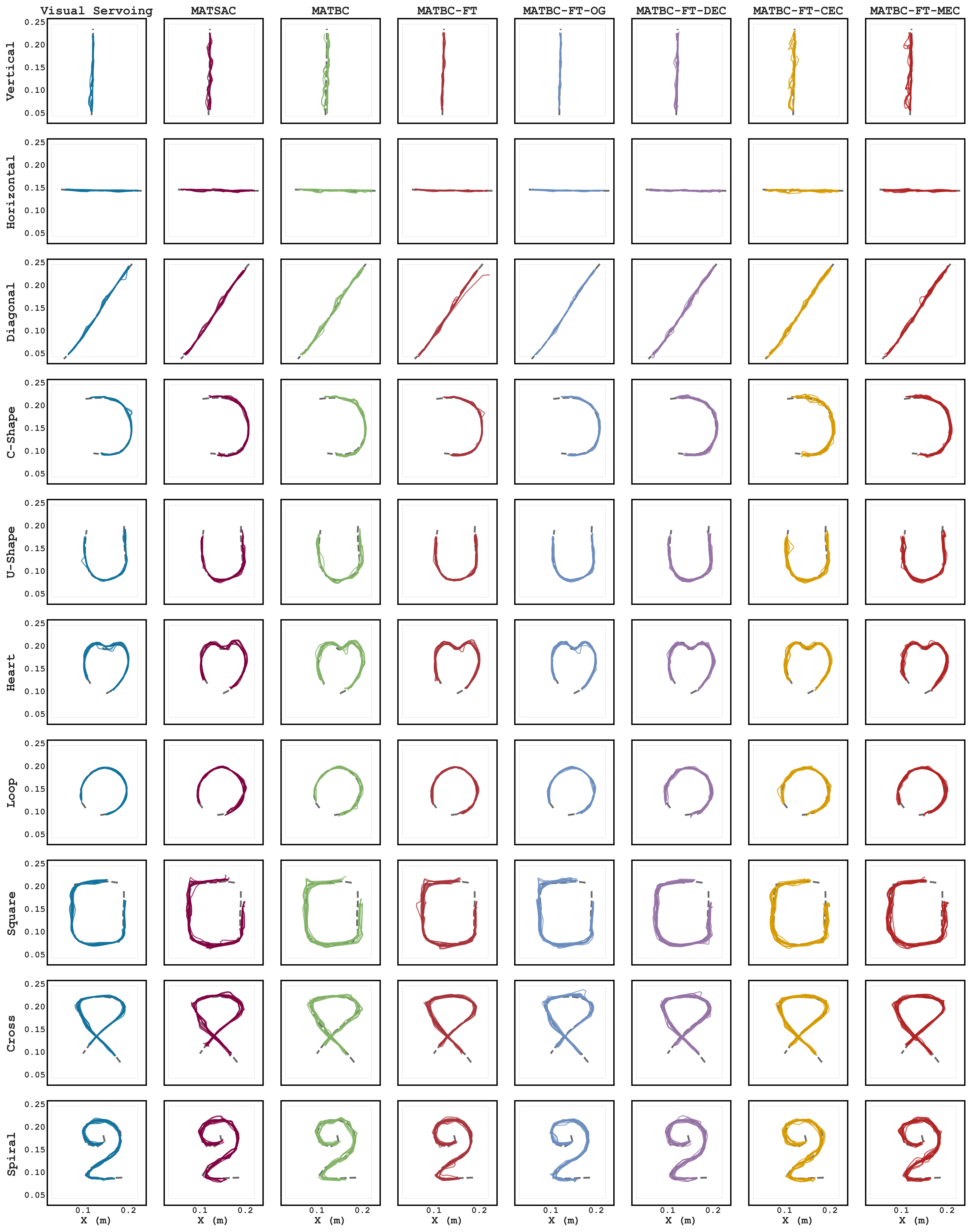}
    \caption{Comparison between different methods while tracking objects being manipulated along all target trajectories. For each subplot, we collect data over all 10 objects and run 5 trials in simulation.}
    \label{fig:test_traj_track_sim}
\end{figure}

\clearpage
\section{Spatial Contrastive Embeddings Ablations:}
\subsection{Locally Constrained RoPE (LC-RoPE)}

We ablate the standard RoPE formulation to incorporate spatial locality constraints based on the physical arrangement of delta robots. We introduce a locality mask $M_{ij}$ that constrains attention to spatially proximate robots within the workspace overlap radius $r_{\text{local}} = 6$ cm after the Qeury and Key matrix multiplication in the RoPE attention mechanism:
\begin{equation}
M_{ij} = \begin{cases} 
1 & \text{if } d_{ij} \leq r_{\text{local}} \\
0 & \text{otherwise}
\end{cases}
\end{equation}

The attention weights are computed with spatial locality constraints to ensure attention is restricted to spatially local neighborhoods:
\begin{equation}
A_{ij} = M_{ij} \cdot \frac{\exp\left(\frac{\mathbf{q}_i^T \mathbf{k}_j}{\sqrt{d_k}}\right)}{\sum_{l \in \mathcal{N}_t} M_{il} \cdot \exp\left(\frac{\mathbf{q}_i^T \mathbf{k}_l}{\sqrt{d_k}}\right)}
\end{equation}

\subsection{Learned Relative Embeddings (LRE)}
This ablation replaces the fixed-angle calculation in RoPE with a learned lookup Embedding table corresponding to the relative neighborhoods with respect to the geometry of the hexagonal arrangement of the delta arrays. We categorize embeddings into a finite set of neighborhood classes. For our work, we consider $K=13$ such classes: the robot's own position, its six immediate neighbors, and its six intermediate-nearest neighbors.

\subsubsection{Learnable Angle Lookup Table}
We define a learnable lookup table (nn.embedding), $\boldsymbol{\Phi}$, which stores the rotation angles for each of these $K$ relative position classes.
\begin{equation}
    \boldsymbol{\Phi} \in \mathbb{R}^{K \times d/2}
\end{equation}
Here, $d$ is the model's embedding dimension. This matrix maps each of the $K$ spatial classes to a vector of $d/2$ angles, one for each 2D subspace, mirroring the structure of RoPE. This lookup table is the core learnable component of LRE.

\subsubsection{Mapping and Angle Retrieval}
For any two robots, $i$ and $j$, located at grid positions $(g_x^i, g_y^i)$ and $(g_x^j, g_y^j)$, we first determine their relative position class, $c_{ij}$:
\begin{equation}
    c_{ij} = \rho\left((g_x^j - g_x^i, g_y^j - g_y^i)\right) \in \{0, 1, \dots, K-1\}
\end{equation}
The function $\rho$ is a deterministic mapping that takes a 2D displacement vector on the hexagonal grid and returns its corresponding integer class index. We then use this index to retrieve the learned angle vector $\boldsymbol{\phi}_{ij}$ from the lookup table:
\begin{equation}
    \boldsymbol{\phi}_{ij} = \boldsymbol{\Phi}[c_{ij}] = [\phi_{ij,1}, \phi_{ij,2}, \dots, \phi_{ij,d/2}]
\end{equation}

\subsubsection{Relative Rotation Matrix Construction}
Using the retrieved angles, we construct a block-diagonal rotation matrix $\mathbf{R}_{ij}$ that represents the relative spatial transformation from robot $i$ to robot $j$.
\begin{equation}
    \mathbf{R}_{ij} = \text{BlockDiag}\left(\mathbf{R}^{(1)}_{ij}, \mathbf{R}^{(2)}_{ij}, \ldots, \mathbf{R}^{(d/2)}_{ij}\right)
\end{equation}
Each 2x2 rotation block $\mathbf{R}^{(k)}_{ij}$ is a standard rotation matrix formulated using the corresponding learned angle $\phi_{ij,k}$:
\begin{equation}
    \mathbf{R}^{(k)}_{ij} = \begin{pmatrix} \cos(\phi_{ij,k}) & -\sin(\phi_{ij,k}) \\ \sin(\phi_{ij,k}) & \cos(\phi_{ij,k}) \end{pmatrix}
\end{equation}
This construction is directly analogous to the rotation matrix used in the original RoPE formulation.

\subsubsection{Application in Self-Attention}
Finally, we incorporate this learned relative rotation into the self-attention mechanism. After projecting the input embeddings $\mathbf{x}_i$ and $\mathbf{x}_j$ into queries and keys, we apply the rotation.
\begin{align}
    \mathbf{q}_i &= \mathbf{W}_q \mathbf{x}_i \\
    \mathbf{k}_j &= \mathbf{W}_k \mathbf{x}_j
\end{align}
The attention score between robots $i$ and $j$ is computed by applying the relative rotation matrix $\mathbf{R}_{ij}$ to the key vector before the dot product with the query vector:
\begin{equation}
    \text{score}(i, j) = \frac{(\mathbf{q}_i)^T (\mathbf{R}_{ij} \mathbf{k}_j)}{\sqrt{d}}
\end{equation}
This formulation ensures that the attention score is modulated by a learned geometric relationship, directly injecting the spatial structure of the hexagonal grid into the self-attention mechanism. By making the rotation angles in $\boldsymbol{\Phi}$ learnable, our model can determine the optimal transformations for each spatial relationship through gradient descent, rather than relying on a fixed mathematical formula that may not perfectly capture the underlying physics of the multi-robot system.

\end{document}